\documentclass[a4paper,fleqn]{cas-dc}

\usepackage[numbers]{natbib}
\usepackage{algorithm}
\usepackage{algpseudocode}
\usepackage{amsmath}
\usepackage{graphics}
\usepackage{epsfig}
\usepackage{multirow}
\usepackage{bm}
\usepackage{wrapfig}
\usepackage{graphicx}
\usepackage{amssymb}
\usepackage{subfigure}

\def\tsc#1{\csdef{#1}{\textsc{\lowercase{#1}}\xspace}}
\tsc{WGM}
\tsc{QE}

\begin{document}
\let\WriteBookmarks\relax
\let\printorcid\relax
\def\floatpagepagefraction{1}
\def\textpagefraction{.001}

\shorttitle{When Large Language Model Meets Optimization}

\shortauthors{Huang et al.}

\title [mode = title]{When Large Language Model Meets Optimization}



%
\author[1]{Sen\ Huang}
\ead{huangsen@scut.edu.cn}
\author[2]{Kaixiang\ Yang}
\ead{yangkx@scut.edu.cn}
\author[3]{Sheng\ Qi}
\ead{qisheng@nudt.edu.cn}
\author[3]{Rui\ Wang}
\cormark[1]
\ead{ruiwangnudt@gmail.com}
\affiliation[1]{{School of Electronic and Information Engineering, South China University of Technology}}
\affiliation[2]{{School of Computer Science and Engineering, South China University of Technology}}
\affiliation[3]{{College of Systems Engineering, National University of Defense Technology}}

\cortext[cor1]{Corresponding author.}

\begin{abstract}
Optimization algorithms and large language models (LLMs) enhance decision-making in dynamic environments by integrating artificial intelligence with traditional techniques. LLMs, with extensive domain knowledge, facilitate intelligent modeling and strategic decision-making in optimization, while optimization algorithms refine LLM architectures and output quality. This synergy offers novel approaches for advancing general AI, addressing both the computational challenges of complex problems and the application of LLMs in practical scenarios. This review outlines the progress and potential of combining LLMs with optimization algorithms, providing insights for future research directions.
\end{abstract}




\maketitle

\section{Introduction}

Optimization algorithms (OA) is becoming increasingly important as a class of heuristic search algorithms in the broad field of artificial intelligence and machine learning~\cite{metropolis1953equation,abdel2021hybrid,singh2022hybridizing}. OA draws on the natural mechanisms of biological evolution, including processes such as natural selection, heredity, mutation, and hybridization, for solving complex optimization problems. These algorithms are widely used in many fields due to their global search capability, low dependence on problem structure, and ease of parallelization.
Optimization algorithms pivotal in diverse fields such as logistics, finance~\cite{lv2024developing}, healthcare~\cite{massim2012efficient}, and artificial intelligence~\cite{chen2024evoprompting}, aim to identify the best solution from available alternatives. They are essential for making decisions efficiently and effectively in an era of rapidly increasing data complexity and volume.
The continuous advancement in optimization techniques has resulted in significant enhancements to algorithmic strategies, each customized to address specific types of problems and operational constraints.
From deterministic methods addressing linear problems to stochastic approaches for global optimization under uncertainty, optimization algorithms hold promise across a broad spectrum of research and practical applications.
With the development of technology, especially when dealing with large-scale, high-dimensional and dynamically changing optimization problems, traditional algorithms often face performance bottlenecks. Evolutionary computing provides effective solutions to these problems with its unique search strategy. In addition, the flexibility and adaptability of evolutionary computation enable it to be combined with a variety of other computational techniques to form hybrid algorithms for further performance enhancement.

In the rapidly evolving field of artificial intelligence, Large Language Models (LLMs) such as GPT (Generative Pre-trained Transformer)~\cite{brown2020language} provide a significant breakthrough with their advanced natural language understanding and generation capabilities. These models have revolutionized applications ranging from automated writing assistants to sophisticated conversational agents.
LLMs have become pivotal in advancing fields like natural language processing, image recognition with their extensive parameters and deep learning capabilities, and machine learning, offering robust solutions for complex data-driven challenges.
LLMs have achieved breakthroughs in traditional NLP tasks like text generation and language translation through extensive data training, and they also show promising potential in the emerging fields of algorithm design and optimization.
Traditional optimization algorithm design, dependent on human expertise~\cite{mirsadeghi2021hybridizing}, is both time-consuming and potentially limited by the experts' knowledge. The advent of large-scale language models, however, has transformed this arena. These models learn extensive algorithmic patterns and strategies, enabling them to devise new algorithms and tailor solutions to specific challenges.

Furthermore, constructing and training large language models require significant computational resources and large datasets~\cite{zhao2023survey}, which escalate research and development costs and constrain the applicability and generalization of LLMs.
Optimization algorithms are crucial in developing LLMs, enabling researchers to efficiently tailor and refine model structures for specific applications.
By enhancing the training process, boosting computational efficiency, and lowering resource consumption, these algorithms facilitate the construction and application of large-scale language models.
These algorithms enhance the models' generalization capabilities and robustness, enabling improved performance amidst real-world complexity and uncertainty.
The goal in designing optimization algorithms for LLMs is to enhance their operational efficiency and reduce resource consumption, without compromising, and possibly improving, model performance.

This review aims to systematically analyze research on developing optimization algorithms with LLMs, and optimizing LLMs with optimization algorithms. It summarizes related research and application scenarios, and explores the diverse aspects of these applications.
Section~\ref{llms} provides a comprehensive review of large language model development, also known as macromodels, tracing their progression from basic predictive models to sophisticated systems that can comprehend and generate human-like text.
Section~\ref{optim} focus to optimization algorithms, concisely tracing their evolution from basic iterative methods to advanced algorithms essential for efficiently scaling AI models.
Section~\ref{llm_optim} examines research that approaches large language models as optimization problems, highlighting innovative methods employed as search operators and in designing optimization algorithms.
Section~\ref{optim_llm} discusses recent advances in optimization algorithms tailored for refining large-scale models. It highlights how these algorithms can enhance design, boost efficiency, and improve the performance of LLMs.
Section~\ref{app} examines the practical applications of integrating optimization algorithms with LLMs, highlighting real-world implementations and their benefits.
Section~\ref{future} is future outlook and research trends, in which summarizes the insights gained from this exploration and provides an outlook on the potential future developments in this exciting intersection of AI research.

In summary, we conducted a comprehensive study on the development and application of optimization algorithms for large models, aiming to provide valuable references and insights for future research.

\section{Large language models}\label{llms}
\par Language has a crucial role in human cognition, enabling communication and expression from early childhood to adulthood \cite{lupyan2016centrality}. Teaching robots to imitate human-like language skills is a difficult task due to their intrinsic lack of cognitive capacity for understanding and expressing language. Computational linguistics aims to close this divide by using advanced Artificial Intelligence (AI) algorithms to enable machines to possess reading, writing, and communication skills that like those of humans \cite{turing2021computing}.

\par The rise of Large Language Models (LLMs) undoubtedly represents an important milestone in the evolution of Natural Language Processing (NLP). These models, such as GPT-3 and GPT-4 of the GPT family \cite{brown2020language}, are built on the Transformer architecture and have up to billions of parameters. They achieve a deep understanding of natural language and generative capabilities by pre-training on massive text datasets. The evolution of LLMs has gone through several notable stages: from the early days of Statistical Language Models (SLMs) and Neuro-Linguistic Models (NLMs), to Pre-Trained Language Models (PLMs), and ultimately to today's large-scale language models. While PLMs like BERT and GPT-2 have achieved remarkable success in NLP tasks, the emergence of LLMs has revolutionized the game in this field \cite{chang2024survey}. Not only can they be adapted to a wide range of tasks through large-scale pre-training, but they are also further optimised through fine-tuning, demonstrating a wide range of potential in application scenarios such as chatbots, search engine optimization and office automation The rise of Large Language Models (LLMs) undoubtedly represents an important milestone in the evolution of Natural Language Processing (NLP). These models, such as GPT-3 and GPT-4 of the GPT family \cite{brown2020language}, are built on the Transformer architecture and have up to billions of parameters. They achieve a deep understanding of natural language and generative capabilities by pre-training on massive text datasets. The evolution of LLMs has gone through several notable stages: from the early days of Statistical Language Models (SLMs) and NLMs, PLMs, and ultimately to today's large-scale language models. While PLMs like BERT and GPT-2 have achieved remarkable success in NLP tasks, the emergence of LLMs has revolutionized the game in this field. Not only can they be adapted to a wide range of tasks through large-scale pre-training, but they are also further optimized through fine-tuning, demonstrating a wide range of potential in application scenarios such as chatbots, search engine optimization, and office automation \cite{zhao2023survey}.

\par The excellence of Large Language Models (LLMs) in the field of Natural Language Processing (NLP) is due to several key components of their design, which together give LLMs powerful language understanding and generation capabilities \cite{zhao2023survey}. First, ``pre-training'' is one of the core processes of LLMs. By pre-training on large-scale textual datasets, LLMs are able to learn the basic structures and patterns of language. These datasets typically contain billions of words covering a wide range of topics and language styles, allowing the models to capture the diversity and complexity of the language. Second,``adaptability'' is another key characteristic of LLMs. After pre-training, LLMs can be further fine-tuned to adapt to specific downstream tasks, such as text classification, sentiment analysis, or machine translation. This adaptability allows LLMs to optimise their performance for specific tasks, leading to better results in various NLP challenges \cite{min2023recent}. In terms of ``applications'', the broad applicability of LLMs is another reason for their popularity. Not only do they perform well in traditional NLP tasks, but they can also be applied to a wider range of domains, such as the development of chatbots, the optimization of search engines, the construction of content recommendation systems, and the development of automated office tools. Finally, ``performance evaluation'' is critical to ensure the reliability and effectiveness of LLMs. Through a series of standardised testing and evaluation protocols, researchers are able to quantify the performance of LLMs and ensure that they work consistently across a range of tasks and conditions. Performance evaluation also includes studies of model bias, fairness, and interpretability, which are key factors in improving model quality and trust.

\par LLMs are becoming a key driver in the field of AI, and their development and application are attracting widespread attention from industry and academia. LLMs represented by ChatGPT and GPT-4 have not only made significant progress in technology but also promoted in-depth discussions on artificial general intelligence (AGI) conceptually \cite{zhao2023survey}. OpenAI's technical article proposes that GPT-4 \cite{achiam2023gpt} may be an early attempt to move towards AGI, which all indicates the critical position of LLMs in the development of AI \cite{bubeck2023sparks}. In the field of Natural Language Processing (NLP), LLMs are becoming a common tool for solving various linguistic tasks, changing the previous research and application paradigm. The Information Retrieval (IR) field is also feeling the winds of change, with traditional search engines facing the challenge of emerging information access methods such as AI chatbots, such as New Bing3, which is an attempt to enhance search results based on LLMs \cite{cao2023comprehensive}. In addition, the field of computer vision (CV) is exploring multimodal models that combine vision and language, and the multimodal input support of GPT-4 is a manifestation of this trend. The rise of LLMs heralds the birth of a whole new ecosystem of apps based on these advanced models. Microsoft 365 leverages LLMs (e.g., Copilot) to automate the office, while OpenAI introduces plug-in functionality in ChatGPT, all of which demonstrate the potential of LLMs to enhance productivity and extend application scenarios \cite{wu2023visual}. 

While LLMs have brought about many positive changes, they also present a number of challenges, particularly in terms of the security and accuracy of the generated content. In addition, the training of LLMs requires substantial computational resources, which is a challenge for research institutes as it limits the ability to perform extensive experiments and optimization of the models \cite{chang2023survey}.We have summarised the relevant restrictions below: 1) {\bf{Computational resources}}: LLMs demand substantial computational resources for training and inference, which can pose challenges for implementing optimization algorithms at scale. Ensuring access to adequate computational infrastructure remains a significant hurdle. 2) {\bf{Data efficiency}}: While LLMs have demonstrated impressive performance, they often require large amounts of data for effective training. This reliance on extensive datasets can be a bottleneck for optimization algorithms, especially in scenarios where data availability is limited or costly to obtain. 3) {\bf{Interpretability and explainability}}: The inherent complexity of LLMs poses challenges for the interpretability and explainability of optimization algorithms. Understanding the decision-making process of these models and interpreting their outputs can be challenging, particularly in critical applications where transparency is essential. 4) {\bf{Generalization and Robustness}}: Ensuring the generalization and robustness of optimization algorithms trained using LLMs is another key challenge. Over-reliance on specific patterns in the training data may lead to poor generalization performance on unseen data or vulnerability to adversarial attacks.

\section{Classic Optimization Algorithm}\label{optim}

\subsection{Traditional optimization algorithms for
optimization problems}
Optimization algorithms have wide applications in fields such as industry, economics, and management. With the rapid development of artificial intelligence, optimization algorithms play a crucial role in achieving intelligence and automation. Traditional optimization algorithms include deterministic optimization algorithms, approximation algorithms, and heuristic algorithms. Deterministic optimization algorithms include  Linear Programming (LP)\cite{newman2013survey}, Integer Programming (IP)\cite{savelsbergh1997branch}, Mixed Integer Programming (MIP)\cite{gleixner2021miplib}, Convex Optimization\cite{Wang2020StochasticGD}, Adaptive Dynamic Pro gramming (ADP)\cite{liu2020adaptive}, etc. Deterministic optimization algorithms can guarantee finding the global optimal solution, but they are generally not suitable for large-scale problems. Polynomial-time approximation algorithms can find a good solution within a reasonable time frame, but they do not guarantee optimality. In other words, approximation algorithms do not guarantee finding the global optimal solution. For some specific problems, optimality guarantees may not exist at all. Heuristic algorithms employ specially designed functions to intelligently explore the solution space \cite{desale2015heuristic}. Heuristic Algorithm rely on intuitive rules, trial-and-error strategies, and practical insights to approximate solutions within acceptable bounds. Heuristic algorithms are particularly effective for problems with high-dimensional search spaces or combinatorial complexities where exact solutions are impractical. Heuristic algorithms include greedy algorithm\cite{cerrone2017carousel}, Tabu Search\cite{alba2006metaheuristic, amuthan2016survey, gupta2020optimizing}, genetic algorithm\cite{wang2020new, lee2021genetic}, differential evolution algorithm\cite{neri2010recent, das2016recent}, Cooperative co-evolution algorithm\cite{shi2005cooperative,cao2015effective,trunfio2016new}. Heuristic algorithms have been widely used due to their excellent computational performance, but they typically require customization and domain expertise for specific problems. Additionally, heuristic algorithms may converge to local optimal solutions and have high time complexity. 

Recently, the superiority of using Estimation of Distribution Algorithm (EDA) in solving optimization problems has been demonstrated. EDA is a prominent optimization technique that employs probabilistic models to guide the search process. Unlike traditional evolutionary algorithms, which rely on mutation and recombination operators, EDA focuses on building and updating a probabilistic model of promising solutions. This model is then sampled to generate new candidate solutions, effectively balancing exploration and exploitation. Yang et al. \cite{yang2021adaptive} proposed ACSEDA based on the Gaussian distribution model, and calculates the covariance according to an enlarged number of promising individuals. In contrast to solely relying on solutions from the current generation to estimate the Gaussian model, $EDA^{2}$ \cite{liang2018enhancing} incorporates a strategy where a set number of high-quality solutions from previous generations are retained in an archive. These historical solutions are then utilized to aid in estimating the covariance matrix of the Gaussian model. Dong et al. \cite{dong2019latent} introduced a latent space-based EDA (LS-EDA), which converts the multivariate probabilistic model of Gaussian-based EDA into a principal component latent subspace with reduced dimensionality. This transformation effectively decreases the complexity of EDA while preserving its probability model, ensuring that crucial information is retained. Consequently, LS-EDA enhances performance scalability for large-scale global optimization problems. In recent times, hybrid EDAs have emerged as a prominent research focus. Li et al. \cite{li2023improved} proposed IDE-EDA, an enhanced version of DE achieved by integrating the EDA. Zhang et al. \cite{zhang2004hybrid} introduced a new hybrid evolutionary algorithm for continuous global optimization problems, Zhou et al. \cite{zhou2015estimation} proposed a fusion of an Estimation of Distribution Algorithm (EDA) with both economical and costly local search (LS) methods. This integration aims to leverage both global statistical insights and individual location-specific information for enhanced optimization performance.

\subsection{Reinforcement learning methods for optimization problems}

Another famous learning paradigm is reinforcement learning. Reinforcement learning learns and improves its strategy by interacting with the environment, aiming to maximize long-term cumulative rewards. The aforementioned heuristic algorithms typically find optimal or approximate solutions by searching the solution space, guided by heuristic information. However, they do not consider interaction with the environment during the search process. Unlike supervised and unsupervised learning, in reinforcement learning, agents can only learn through trial and error, rather than relying on labeled data or searching for the inherent structure of the data. Furthermore, reinforcement learning can be categorized into classical reinforcement learning methods and deep reinforcement learning methods, which will be introduced below.

According to \cite{lei2022solve}, classic RL methods can be divided into model-based and model-free approaches. Model-free methods can be further categorized into value-based, policy-based, and actor-critic methods. Value-based methods seek the optimal policy by estimating the value function. This approach is suitable for smaller state spaces and discrete action spaces but faces challenges to extend to continuous action spaces and has the "high bias" problem. The error between the estimated value function and the actual value function is difficult to eliminate. Policy-based methods, on the other hand, do not require estimating the value function. Instead, they directly fit the policy function using neural networks. By training and updating the policy parameters, the optimal policy is generated. This method is suitable for continuous action spaces but requires sampling a large number of trajectories, and there is a huge difference between each trajectory, leading to the "high variance" problem. To address the contradiction between high bias and high variance, the actor-critic method emerges. The actor-critic method constructs an agent that can both output policies and evaluate their quality in real-time using the value function. Generally, an actor-critic network consists of two parts: the actor network and the critic network. The actor network is used to generate policies to approximate the policy function, while the critic network is used to evaluate policies to approximate the value function. Representative works of these methods will be introduced below.

Q-learning\cite{watkins1989learning} is a classic value-based RL algorithm and is currently the most widely used model-free RL algorithm. Q-learning first initializes a Q-function, typically represented as a Q-table. It selects an action based on the current state using the $\epsilon$-greedy strategy, performs the selected action, and observes the reward obtained and the next state transitioned to. The Q-function is updated using the Bellman equation, repeat the above steps and update the Q value until the stop condition is reached. Finally, the optimal policy is extracted based on the learned Q-function. Double Q-learning\cite{hasselt2010double} is an improved version of the Q-learning algorithm which alleviates the overestimation problem by using two Q-functions. In each round of interaction with the environment, one of the value function estimators is alternately selected to choose actions, while the other is used for action value estimation. Existing research has demonstrated that the Double Q-learning method achieves higher stability and greater long-term returns.

The REINFORCE algorithm\cite{williams1992simple}, also known as the Monte Carlo Policy Gradient Reinforcement Learning algorithm, is a classical policy gradient method. The goal of the REINFORCE is to update parameters along the gradient direction to maximize the objective function. On the basis of the classical policy gradient algorithms, Trust Region Policy Optimization (TRPO)\cite{schulman2015trust} ensures that the Kullback-Leibler Divergence between the new and the old policy does not exceed a predefined threshold during each policy update. This threshold represents the "trust region" of policy updates, indicating the similarity between the new and old policies, thereby ensuring the stability of policy optimization. Proximal Policy Optimization (PPO)\cite{schulman2017proximal} is an improvement over TRPO, which is simpler to implement and requires less computation in practical use.

Actor-Critic (AC)\cite{konda1999actor} algorithm combines the advantages of both policy-based and value-based methods. It learns both the policy and the value function simultaneously. The actor trains the strategy based on the value function feedback from the critic, while the critic trains the value function using the Time Difference Method (TD) for single step updates. The aforementioned REINFORCE algorithm uses a stochastic policy function, which outputs the probability distribution of actions for a given state, and then selects an action based on the probability distribution. In contrast to the REINFORCE algorithm, Deterministic Policy Gradient (DPG)\cite{silver2014deterministic} algorithm employs deterministic policy function, which directly outputs a deterministic action for given states and update policy parameters by maximizing expected returns. DPG integrates deterministic policy gradients only in the state space, greatly reducing the need for sampling and enabling the handling of larger action spaces. However, the coupling between policy updates and value estimation in DPG leads to insufficient stability, particularly being highly sensitive to hyperparameters. The difficulty of tuning hyperparameters in actor-critic algorithms and challenges in reproducibility make them hard to apply in practical scenarios. When extended to application fields, the robustness of the algorithm is also one of the most concerned core issues.

The above methods are relatively simple and intuitive, easy to understand and implement, and suitable for smaller-scale problems. In relatively fewer samples, classical RL methods typically achieve good performance, especially in stable environments and reward settings. More importantly, the above method exhibits strong interpretability, allowing for a clear understanding of the agent's behavior and decision-making process within the environment.

Although RL has achieved remarkable achievements, previous methods have often struggled to handle high-dimensional data such as images, text, etc. This limitation has constrained its ability to deal with complex tasks and environments. Classical RL methods often find it challenging to strike a good balance between exploration and exploitation, leading to susceptibility to local optima, especially in high-dimensional and complex environments where issues of insufficient exploration are more pronounced. The reason for the aforementioned situation is that RL algorithms, like other algorithms, face challenges such as memory complexity, computational complexity, sample complexity, etc\cite{arulkumaran2017deep}. However, the powerful representation learning capability and function approximation capability of deep learning bring a completely new solution for RL.

Deep learning is a branch of machine learning aimed at using multi-layer neural network models to learn representations and features of data, and solving various tasks through these representations and features. Deep learning models have strong nonlinear function approximation capabilities, allowing them to learn more complex and accurate data representations and features. Deep learning models have strong generalization and representation learning capabilities, enabling them to learn more accurate and effective policies or value functions, thereby allowing RL agents to tackle more complex and high-dimensional tasks and environments, achieving higher performance and accuracy. Deep reinforcement learning (DRL) is the product of integrating both RL and DL. Similarly, DRL methods are also divided into three types: value-based, policy-based, and actor-critic methods. Among them, value-based DRL employs DL to approximate value functions, while policy-based DRL uses DL to approximate policies and solve decision-making policies based on policy gradient rules. The following will introduce representative works in DRL.

In 2013, Mnih et al. from DeepMind combined DL with Q-learning, proposing the groundbreaking Deep Q-network (DQN)\cite{mnih2013playing}. DQN is a DRL algorithm based on Q-learning. On one hand, it utilizes deep neural networks as value function estimators, and on the other hand, it introduces experience replay and target networks. The experience replay mechanism breaks the high dependency between sampled samples, while the target network alleviates the instability of neural networks during training. These two mechanisms work together to enable the DQN algorithm to achieve performance close to or even surpassing human levels in most Atari games. Double DQN\cite{van2016deep}, based on Double Q-learning, is an improvement over DQN. Similar to how DQN extends Q-learning, DDQN addresses the overestimation issue by using two Q-networks. DDQN achieves better stability and algorithm performance compared to DQN. In 2017, Dai et al. combined RL with graph embedding and proposed S2V-DQN\cite{khalil2017learning}. They utilized a graph embedding network called structure2vec (S2V) to represent the policy in greedy algorithms and employed multi-step DQN to learn greedy policies parameterized by the graph embedding network. The S2V-DQN algorithm generates high-quality solutions faster, sometimes finding better solutions than commercial solvers within a longer timeframe.

In 2015, inspired by the ideas of DQN, Lillicrap et al. combined neural networks with the DPG algorithm to propose DDPG\cite{lillicrap2015continuous}. DDPG employs two different parameterized deep neural networks to represent the value network and the deterministic policy network. The policy network is responsible for updating the policy, while the value network outputs the Q-values for state-action pairs. Similar to DQN, DDPG also utilizes target networks to overcome the instability issues during network updates. Zhang et al. \cite{zhang2020deep} considered the feasibility constraints of NP-hard problems, embedding heuristic functions into the transformer architecture, and applying DRL to combinatorial optimization problems with feasibility constraints. Ma et al.\cite{ma2019combinatorial} introduced a Graph Pointer Network (GPN) to solve the classical TSP problem and combined it with a hierarchical RL framework to address the TSP problem with time window constraints. Multiple Traveling Salesman Problems (MTSP) are more complex, and Hu et al.\cite{hu2020reinforcement} designed a network consisting of a shared graph neural network and distributed policies to learn a common policy expression suitable for MTSP. Experimental results demonstrated the effectiveness of this approach on large-scale problems.

In 2016, Mnih et al.\cite{mnih2016asynchronous} developed an improved Actor-Critic algorithm called A3C (Asynchronous Advantage Actor-Critic). By utilizing asynchronous gradient descent to optimize the parameters of deep neural networks (DNNs), A3C significantly improved the efficiency of policy optimization. Vinyals et al. \cite{vinyals2015pointer} proposed the Pointer Network model for solving combinatorial optimization problems, which initiated a series of research studies on utilizing DNN for solving combinatorial optimization problems. This model was inspired by the Seq2Seq model in machine translation. It employs a deep neural network-based encoder to encode the input sequence of the combinatorial optimization problem (such as city coordinates), then utilizes a decoder and attention mechanism to compute the selection probabilities of each node. Finally, it selects nodes in an autoregressive manner until obtaining a complete solution. Due to the supervised nature of the training method proposed by Vinyals et al.\cite{vinyals2015pointer}, the quality of the solutions it obtains will never exceed the quality of the sample solutions. Recognizing this limitation, Bello et al.\cite{bello2016neural} employed a RL approach to train the Pointer Network model. They treated each problem instance as a training sample, using the objective function of the problem as feedback signals, and trained the model based on REINFORCE. They also introduced a critic network as a baseline to reduce training variance. Furthermore, Nazari et al.\cite{nazari2018reinforcement} extended the Pointer Network to handle dynamic VRP problems. They replaced the LSTM in the Encoder input layer with a simple one-dimensional convolutional layer, effectively reducing computational costs. While maintaining optimization effectiveness, the training time was reduced by 60\%.

In recent years, the Transformer\cite{vaswani2017attention} has achieved tremendous success in the field of natural language processing. Its multi-head attention mechanism enables better extraction of deep features from problems. In view of this, several recent studies have drawn inspiration from the Transformer for solving combinatorial optimization problems. Deudon et al. \cite{deudon2018learning} improved traditional pointer network models by incorporating ideas from the Transformer. They utilized a similar structure to the Transformer in the encoder, while the decoder employed linear mapping of the decisions from the last three steps to obtain a reference vector, thereby reducing model complexity. The attention calculation method remained the same as in traditional Pointer Network models, and the classic REINFORCE method is still used to train the model. Kool et al.\cite{kool2018attention} proposed a new method capable of solving multiple combinatorial optimization problems using attention mechanisms. The attention calculation method in this model adopted the self-attention computation method from the transformer, with additional computational layers to enhance performance. They further designed a greedy rollout baseline to replace the Critic network, leading to significant improvements in optimization performance.

Deep reinforcement learning, as one of the most popular research directions in the field of artificial intelligence, has shown great potential in solving complex tasks and addressing various real-world problems. However, DRL also has its limitations, such as data requirements, sample efficiency, computational resources, interpretability, etc. Despite achieving some success in both research and application domains, DRL fundamentally remains constrained to simulated environments with ideal, highly structured experimental data design. They heavily rely on the design and training of specific models. Therefore, there is a growing interest in the design and optimization of automatic algorithms.

\section{LLMs as Optimization}\label{llm_optim}

\subsection{LLMs as the Black-box Optimization Search Model}
\begin{figure}[ht]
  \centering
  \centerline{\includegraphics[width=0.95\linewidth]{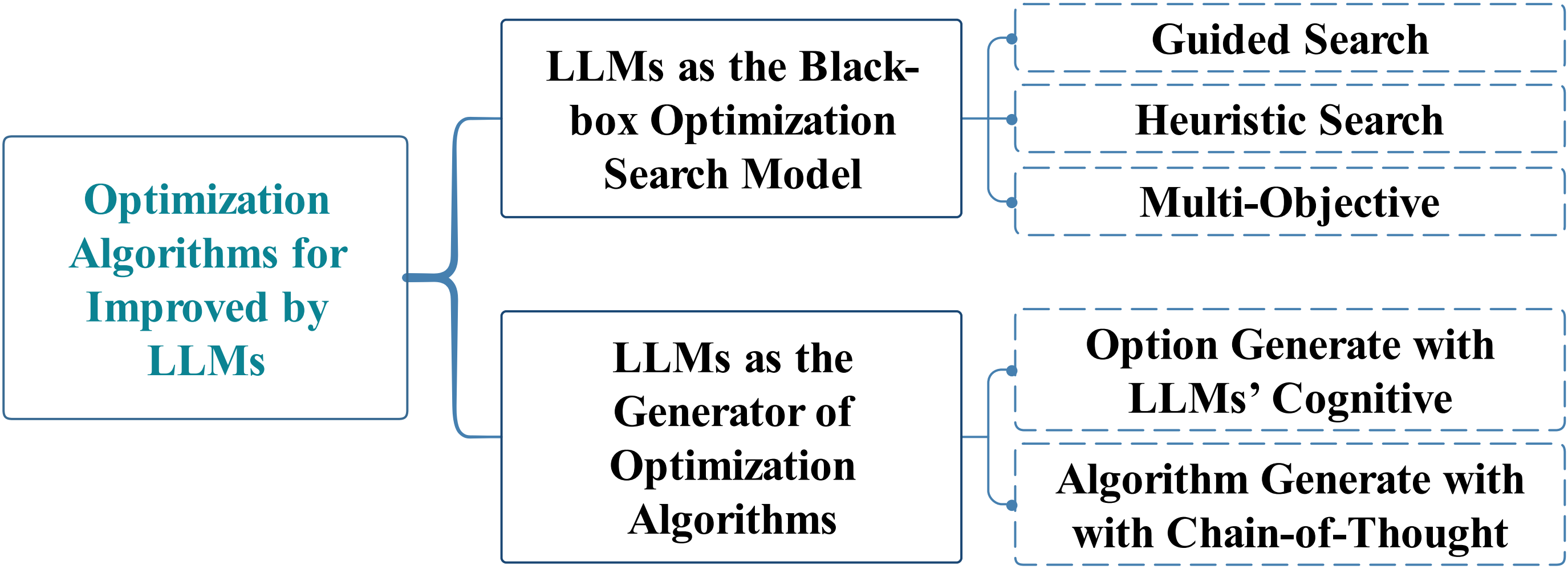}}
\caption{The split of LLMs assist optimization algorithms.}
\label{lao_fig}
\end{figure}

There is a strong alignment between Large Language Models (LLMs), which are powerful in generating creative texts, and Evolutionary Algorithms (EAs), capable of discovering diverse solutions to complex real-world problems~\cite{chao2024match}.
With their powerful knowledge storage and generation capabilities, LLMs can support optimization algorithms in problem decomposition, parameter search, and solution generation.
As show in Fig. \ref{lao_fig}, LLMs can be classified into two categories for enhancing optimization algorithms: 1) One category is to use the large model as the search operator of the black-box optimization model, which makes full use of the knowledge storage capacity and experience of LLMs  and thus can effectively reduce the input of workforce. 2) The other category of approach is to give full play to the generative capacity of the large model and to make full use of the understanding of the optimization problem by the large model as the input of model  and then generate suitable optimization algorithm configurations or generate optimization algorithms for solving specific problems. The use of large models to assist the design of optimization algorithms has achieved preliminary research and widespread attention. How to give full play to the advantages of large models in the design of algorithms and integrate them into the optimization algorithm framework has become the key to the research in this field. A detailed classification of what is the goal of the auxiliary optimization model with LLM is shown in Table~\ref{split}.

\begin{table*}[tbh]
\caption{The detailed classification of what is the goal of the auxiliary optimization model with LLM}
\centering
\begin{tabular} {p{4.0cm} p{3.8cm} p{5.5cm} p{2.3cm}}
\toprule[1.5pt]
\textbf{Research Field} & \textbf{Category} & \textbf{LLM's Goal} & \textbf{Related Works} \\ \hline
\multirow{3}{4.0cm}[-1.3cm]{\centering LLMs as the Black-box Optimization Search Model}
& \centering Guided Search Operator & LLMs combined with natural language descriptions or manual a priori knowledge search for suitable optimization algorithms & \cite{yang2023large,chen2024prompt,zhang2023automl,ahmaditeshnizi2023optimus}\\
    \cline{2-4}
& \centering Heuristic Search Operators & LLMs rely on knowledge storage and problem analysis capabilities to search for suitable optimization algorithms &\cite{ye2024reevo,zhong2024leveraging,meyerson2023language,liu2023large}\\
    \cline{2-4}
& \centering Multi-Objective Optimization & LLMs search for multiple objectives and trade-offs between the importances of the inter-objectives &\cite{brahmachary2024large,liu2023multilarge,bradley2023quality}\\
    \hline
\multirow{2}{4.0cm}[-0.8cm]{\centering LLMs as the Generator of Optimization Algorithms} & \centering Option Generate with the Cognitive of LLMs & LLMs analyze the problem and generate suitable optimization algorithm options based on a priori optimization algorithm knowledge &\cite{zhang2023using,liu2024large,ma2024large}\\
    \cline{2-4}
& \centering Algorithm Generate with with Chain-of-Thought & Generate optimization algorithms, optimization steps, or code with the help of the chain of thought ability of LLM to solve complex problems &\cite{liu2023algorithm,liu2024example,fernando2023promptbreeder,pluhacek2023leveraging,bradley2024openelm,hemberg2024evolving}  \\
\toprule[1.5pt]
\end{tabular}\label{split}
\end{table*}

Yang et al.~\cite{yang2023large} proposed the Optimization by PROmpting (OPRO), which utilizes natural language descriptions to guide LLMs in searching for solutions for optimization problems. This method is particularly effective for derivative-free optimization scenarios that are common in real-world applications.
Technically, OPRO includes previously generated solutions and their values, allowing the LLM to iteratively improve upon solutions. The framework also intelligently balances benefits and costs, crucial for effective optimization, by adjusting the sampling temperature of the LLM to encourage both refinement of existing solutions and exploration of new ones.
Chen et al.~\cite{chen2024prompt} explored cue optimization methods in multi-step tasks to improve task execution efficiency. By constructing a discrete LLM-based prompt optimization framework, the framework automatically provides suggestions for improvement by integrating human-designed feedback rules and preference alignment. It is noted that while LLM performs well in single-step tasks, real-world multi-step tasks pose new challenges, such as more complex cueing content, difficulty in evaluating individual step impacts, and the fact that different people may have different task execution preferences. To address these issues, the researchers introduced human feedback, leveraging human expertise in providing input and combining it with a genetic algorithm-style framework to optimize cues.
For the current emerging optimization problem of neural architecture search, Zhang et al.~\cite{zhang2023automl} designed an automated machine learning (AutoML) system based on large-scale language models (LLMs), AutoML-GPT. AutoML-GPT utilizes a GPT model as a bridge to connect multiple AI models and dynamically uses optimized hyperparameters to train the models. The system automatically generates corresponding prompt paragraphs to search for the optimal model architecture and parameters by dynamically taking inputs from user requests and data cards. It then automatically executes the entire experimental process, from data processing to model architecture, hyperparameter tuning, and prediction of training logs.
AhmadiTeshnizi et al.~\cite{ahmaditeshnizi2023optimus} proposed a large language model (LLM)-based agent called OptiMUS. OptiMUS is designed to formulate and solve mixed-integer linear programming (MILP) problems from natural language descriptions. It is capable of developing mathematical models, writing and debugging solver code, developing tests, and checking the validity of the generated solutions.
\subsubsection{Heuristic Search Operators} 
To tackle NP-hard combinatorial optimization problems (COPs) using large language models (LLMs) as Hyper-Heuristics (LHHs), Ye et al.~\cite{ye2024reevo} proposed ReEvo, a framework that emulates the reflective design approach of human experts. It leverages the scalable inference capabilities of LLMs, Internet-scale domain knowledge, and powerful evolutionary search strategies. ReEvo operates by generating heuristics through LLMs with minimal human intervention, offering an open-ended heuristic space and the potential for Knowledge and competence beyond that of human experts.
The research aims to address the complexity and heterogeneity of COPs by automating the design process of heuristics, which traditionally requires extensive trial and error from domain experts.
ReEvo incorporates a dual-level reflection mechanism, where short-term reflections are used to analyze heuristics' relative performance, and long-term reflections are accumulated to guide their evolution. This reflective process allows ReEvo to adapt and improve heuristics over time, leading to smoother fitness landscapes and more effective search results.
Zhong et al.~\cite{zhong2024leveraging} introduced a groundbreaking approach to the design of metaheuristic algorithms by utilizing the capabilities of the large language model (LLM) ChatGPT-3.5. They proposed a animal-inspired metaheuristic algorithm named Zoological Search Optimization (ZSO), which is developed to tackle continuous optimization problems. The ZSO algorithm is designed to mimic the collective behaviors of animals, incorporating two key search operators: the prey-predator interaction operator and the social flocking operator, which together balance exploration and exploitation effectively.
A similar approach called Language Model Crossover (LMX) utilizes large pre-trained Language Models (LLMs) to generate new candidate solutions.~\cite{meyerson2023language} LMX does this by combining the parent solutions into a prompt and then feeding that prompt into the LLM to collect offspring from the output. This approach is simple to implement and generates high-quality progeny across various domains, including binary strings, mathematical expressions, English sentences, image generation prompts, and Python code.
Liu et al.~\cite{liu2023large} explore the potential of using Large Language Models (LLMs), such as GPT-4, to generate novel hybrid population intelligence optimization algorithms. The research focuses on using GPT-4 to identify and decompose six population algorithms that perform well in sequential optimization: particle swarm optimization (PSO), cuckoo search (CS), artificial bee colony algorithm (ABC), grey wolf optimizer (GWO), self-organizing migration algorithm (SOMA), and whale optimization algorithm (WOA) by constructing hints to guide the LLMs to search for the optimal from the current population's parent solution.
INSTINCT (INSTruction optimization using Neural bandits Coupled with Transformers)~\cite{lin2023use} is a black-box LLMs cue optimization method. The method employs a novel neural band algorithm, Neural Upper Confidence Bound (NeuralUCB), in place of the Gaussian Process (GP) model in BO.NeuralUCB uses a neural network (NN) as a proxy while retaining the theoretical basis of the trade-off between exploration and exploitation in BO. Theoretical basis for the trade-off between exploration and exploitation. More importantly, NeuralUCB allows for the natural coupling of NN agents with hidden representations learned from pre-trained transformers (i.e., open-source LLMs), significantly improving algorithmic performance.

\subsubsection{Multi-Objective Optimization} 
Brahmachary et al.~\cite{brahmachary2024large} introduces an approach to numerical optimization using Large Language Models (LLMs) called Language-Model-Based Evolutionary Optimizer (LEO), which leverages the reasoning capabilities of LLMs to perform zero-shot optimization across a variety of scenarios, including multi-objective and high-dimensional problems.
LEO lies in its population-based strategy, which incorporates an elitist framework consisting of separate explore and exploit pools of solutions. This strategy not only harnesses the optimization capabilities of LLMs but also mitigates the risk of getting stuck in local optima. The method is distinct from other auto-regressive, evolutionary, or population-based methods as it uses LLMs to generate new candidate solutions, providing a unique balance between exploration and exploitation. It is shown through a series of test cases that LEO is not only capable of handling single-objective but also of solving multi-objective optimization problems well.
Liu et al.~\cite{liu2023multilarge} leverage the capabilities of large language models (LLMs) to design operators for multi-objective evolutionary algorithms (MOEAs). The research addresses the challenges associated with the manual design of search operators in MOEAs, which often require extensive domain knowledge and can be time-consuming. The authors propose a method for decomposing the multi-objective optimization problem (MOP) into several single-objective subproblems (SOPs). LLMs are employed as search operators for each subproblem through prompt engineering. This allows the LLM to serve as a black-box search operator in a zero-shot manner without problem-specific training.
Bradley et al.~\cite{bradley2023quality} introduce a new approach called Quality-Diversity through AI Feedback (QDAIF) that combines evolutionary algorithms and large-scale language models (LLMs) to generate high-quality and diverse candidates for optimization algorithms. The core idea of QDAIF is to use language models to create variants and evaluate the quality and diversity of the candidates. The EA is responsible for maintaining the library of optimization algorithms and replacing the newly generated higher quality and more diverse solutions to the relevant positions in the library based on the evaluation of the LLMs to achieve an iterative optimization search process.QDAIF can find a set of higher quality and more diverse solutions within the search space, which is a successful application of the LLMs in QD problems.

\subsection{LLMs as the Generator of Optimization Algorithms}
Optimization algorithms usually require the design of suitable optimization schemes based on the specified tasks. Due to the problem-understanding and algorithmic analysis capabilities of LLMs, the design of optimization methods based on LLMs can generate suitable optimization method selection and combination schemes compared to the optimization methods in the pre-LLM era~\cite{tian2022local, tian2017indicator}.

\subsubsection{Option Generate with the Cognitive of LLMs}

Zhang et al.~\cite{zhang2023using} explore the use of large language models (LLMs) to generate optimal configurations during Hyperparametric Optimization (HPO). The generation method does not rely on a predefined search space and consists of selecting parameters that can be optimized and specifying bounds for these parameters. Furthermore, the process treats the code of the specified model as hyperparameters to be output by the LLM, which goes beyond the capabilities of existing HPO methods.
Liu et al.~\cite{liu2024large} proposes a system named AgentHPO, which leverages the advanced capabilities of LLMs to streamline the HPO process, traditionally a labor-intensive and complex task that requires significant computational resources and expert knowledge. AgentHPO lies in its unique architecture that incorporates two specialized agents: the Creator and the Executor. The Creator agent interprets task-specific details provided in natural language and generates initial hyperparameters (HPs), emulating the role of a human expert. It utilizes extensive domain knowledge and sophisticated reasoning to propose HP configurations that are expected to yield optimal model performance.
Ma et al.~\cite{ma2024large} explored the effectiveness of Large Language Models (LLMs) as cueing optimizers. They find that LLM optimizers struggle to accurately identify the root cause of errors during reflection and often fail to generate appropriate cues for the target model through a single cueing optimization step, even when semantically valid. Further, they proposed a new paradigm of "automated behavioral optimization" designed to more controllably and directly optimize the behavior of the target model.

\subsubsection{Algorithm Generate with with Chain-of-Thought}

Liu et al.~\cite{liu2023algorithm} proposed a approach called Algorithmic Evolution using Large Language Models (AEL), which aims to automate the generation of efficient algorithms for specific optimization problems.AEL creates and improves algorithms by interacting with Large Language Models (LLMs) within an evolutionary framework, eliminating the need for model training and significantly reducing the need for domain knowledge and expert skills. The constructive algorithms generated by AEL outperform hand-crafted heuristics and algorithms generated directly from LLMs based on the Traveling Provider Problem (TSP) example. Further, they introduce a improved framework for automated algorithm design that combines evolutionary computation and LLMs~\cite{liu2024example}. By automating algorithm design, combination, and modification, the AEL framework significantly reduces manual effort and eliminates the need for model training. The researchers used the AEL framework to design a Guided Local Search (GLS) algorithm for solving the TSP.
The PromptBreeder (PB) system~\cite{fernando2023promptbreeder} utilizes the Chain-of-Thought Prompting strategy, which can significantly improve the reasoning ability of Large Language Models (LLMs) in different domains. It is a generalized mechanism for self-referential self-improvement of large language models (LLMs). At the heart of the system is self-reference: not only do the task prompts evolve, but the mutation prompts used to generate them also improve over time.PB outperforms existing state-of-the-art prompting strategies, such as Chain-of-Thought and Plan-and-Solve prompts, in several commonly used benchmark tests—and-Solve prompts.
Pluhacek et al.~\cite{pluhacek2023leveraging} identified and decomposed six population algorithms that perform well on sequential optimization problems through LLM by augmenting the population. Enhanced Swarm Exploration and Exploitation Optimizer(ESEEO) aims to maintain population diversity and effectively balance exploration and exploitation by combining elements of Particle Swarm Optimization (PSO), Cuckoo Search (CS), and Artificial Bee Colony (ABC). Combining Limited Evaluation Population Optimizer (LESO) Designed to solve expensive optimization problems with a limited number of objective function evaluations, LESO combines the features of PSO, Grey Wolf Optimizer (GWO), and ABC to achieve effective exploration and exploitation within a limited number of assessments.
LLMs are a challenge in Genetic Programming (GP) approaches. Bradley et al.~\cite{bradley2024openelm} present an optimized algorithm generation tool for implementing LLMs that converts natural language descriptions into implementation code and automatically repairs program errors. In addition, this paper presents two uses of LLMs as evolutionary operators: difference modeling and LMX crossover. The former is a language model specialized for predicting code discrepancies, and the latter is a technique for generating candidate solutions using multiple parents.
Similarly, Hemberg et al.~\cite{hemberg2024evolving} proposed a LLM-based GP algorithm, called LLM-GP, for generating optimization algorithm code.
Unlike conventional GP algorithms, LLM-GP harnesses the power of pre-trained pattern matching and sequence complementation capabilities of the LLM. This unique feature allows for the design and implementation of genetic operators, paving the way for more efficient and effective algorithm generation.

\section{Optimization algorithms optimize large language models}\label{optim_llm}
\par
\begin{figure*}
    \centering
    \includegraphics[width=0.85\linewidth]{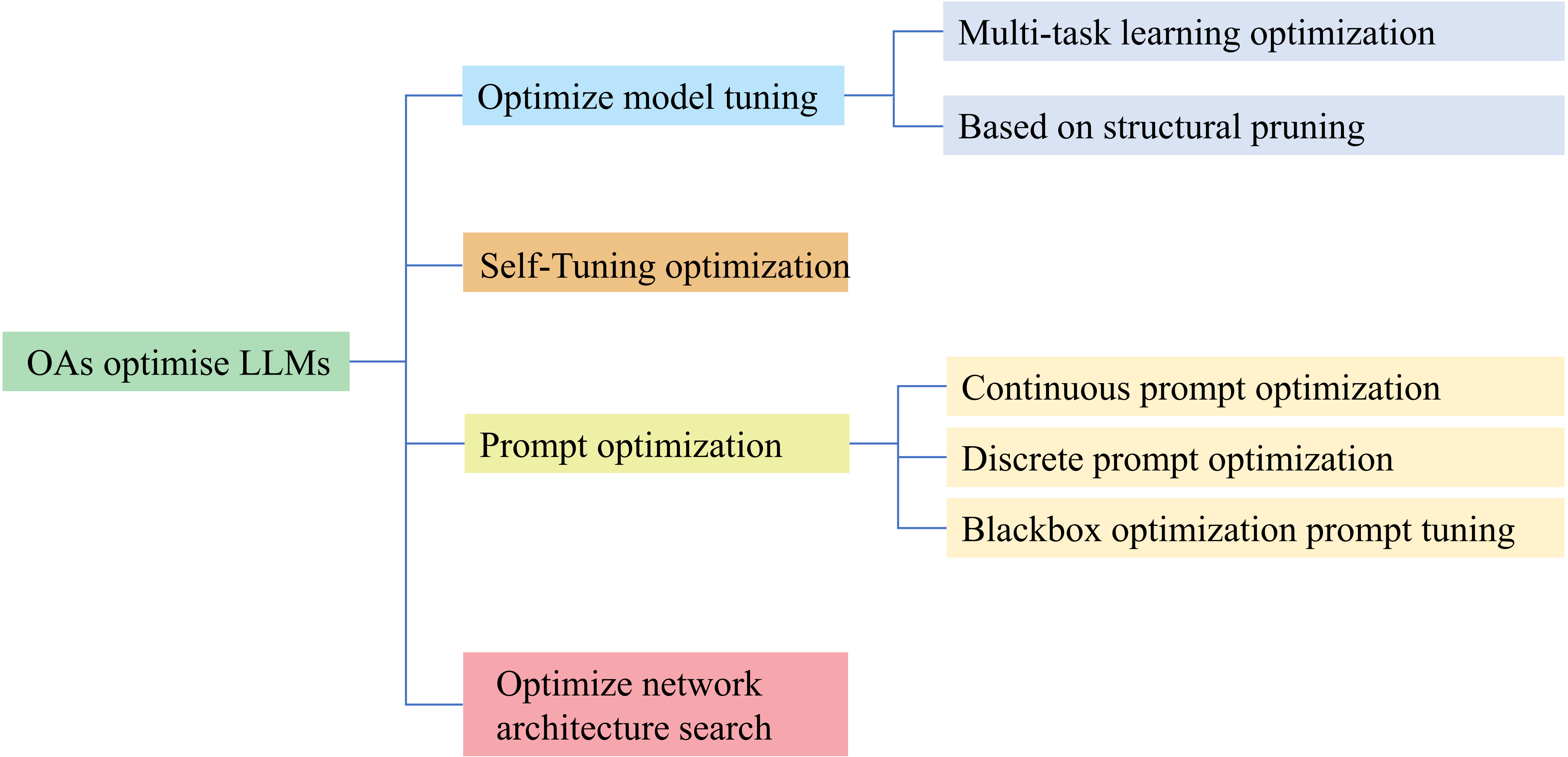}
    \caption{OAs methodological framework for optimizing LLMs}
    \label{fig:fig1}
\end{figure*}

 Large Language Models (LLMs) have exhibited exceptional proficiency in many natural language processing tasks, encompassing text generation and sentiment analysis. Nevertheless, the task of optimizing their performance and efficiency continues to be a crucial obstacle, especially in intricate and unpredictable circumstances. Optimization Algorithms (OA) have emerged as a promising method to improve LLMs. OA utilizes the concepts of natural selection to perform repeated searches inside the parameter space of LLMs \cite{chao2024match, wu2024evolutionary}. It focuses on tasks like prompt engineering, model architecture optimization, hyperparameter setting, and multi-task learning. This holistic strategy seeks to discover optimal solutions for problems that have extensive search spaces, thereby enhancing the effectiveness of LLMs in several fields, such as natural language processing, software engineering, and neural architecture search. In this section, we will delve into the complexities of OA optimization for LLMs, exploring various strategies such as model tuning, prompt tuning and network architecture search to unlock the full potential of these transformative language models. The framework of the optimization algorithm to optimize LLMs is shown in Fig. \ref{fig:fig1}.

\subsection{Optimize model tuning}

\begin{table*}[tbh]
\caption{Summary of model tuning based on OAs} \label{model-tuning}
\centering
\begin{tabular} {p{3.0cm} p{6.4cm} p{3.2cm} p{3.0cm}}
\toprule[1.5pt]
\textbf{Algorithms} & \textbf{Main tasks}                                                                      & \textbf{0bjectives}                            & \textbf{Using OAs}             \\ \hline
Multi-task learning \cite{choong2023jack} & JATs Implementation of One-off Diverse Compact Set of Sets Machine Learning Models       & Multi-tasking and multi-objective optimization & Neuroevolutionary multitasking \\ \hline
EMO-Prompts \cite{baumann2024evolutionary}         & Generate prompts that lead LLM to produce text with two conflicting emotions             & Evolutionary Multi-Objective                   & NSGA-II                        \\ \hline
Structured Pruning \cite{klein2023structural}  & Subparts of the network with optimal performance after fine-tuning for model compression & Multi-objective NAS                            & Weight Shared NAS              \\ \hline
LLM-Pruner \cite{ma2023llm}          & Model compression in a task agnostic manner                                              & Single-objective                               & Low-Rank Adaptation            \\ 
\bottomrule[1.5pt]
\end{tabular}
\end{table*}

\subsubsection{Multi-task learning optimization}
\par Multi-task learning (ML) optimization of large models is an approach that uses optimization methods such as evolutionary algorithms to optimize model architectures for multiple tasks simultaneously. Through multi-task learning, researchers identify optimal architectures for multiple target tasks simultaneously in large pre-trained models \cite{chao2024match,liu2023pre}. The advantage of this approach is the ability to share and interactively learn relevant information between different tasks, thus improving the generalization ability and efficiency of the model. Table \ref{model-tuning} summarises the main optimization model tuning class methods

\par The optimization strategy of multi-task learning reduces training costs, enhances model performance and improves adaptability across diverse domains and tasks by concurrently identifying optimal model architectures for multiple objectives using evolutionary algorithms \cite{wei2021review}. For example, Choong et al. \cite{choong2023jack} proposed the concept of a diverse set of compact machine learning model sets designed to efficiently address multiple target architectures in large pre-trained models through an evolved multi-task learning paradigm. Baumann et al. \cite{baumann2024evolutionary} present an evolutionary multi-objective approach designed to optimize prompts in large language models (e.g., ChatGPT), demonstrating its effectiveness in crafting prompts in a sentiment analysis task that effectively captures conflicting emotions. Yang et al. \cite{yang2023instoptima} treated instruction generation as an evolutionary multi-objective optimization problem, using a large-scale language model to simulate instruction operators in order to improve the quality and diversity of generated instructions.

\par Meanwhile, Gupta and Bali et al. \cite{bali2020cognizant,gupta2016multiobjective} have investigated the use of multi-objective multi-task evolutionary algorithms, such as the MO-MFEA algorithm, in creating task-specific, small-scale models derived from Large Language Models (LLMs) in the field of online search architecture study. These specialized models are created from the LLM as a general base model and exhibit enhanced performance or more compression in different application fields and neural network designs.

\subsubsection{Based on structural pruning}
\par Structural pruning optimizes large language models (LLMs) by selectively removing non-critical coupled structures based on gradient information, effectively reducing model size while preserving functionality and ensuring task-agnosticism. Structural pruning is an essential optimization technique used to enhance pre-trained LLMs for subsequent tasks, such as text categorization and sentiment analysis. Structural pruning, as suggested by Klein \cite{klein2023structural}, seeks to uncover numerous subnetworks of LLMs that achieve a compromise between performance and size, making them easier to use in different real-world applications. This approach employs a multi-objective local search algorithm to identify numerous Pareto-optimal subnetworks effectively. It does this by minimizing evaluation costs via weight sharing. Ma et al. \cite{ma2023llm} propose the LLM-Pruner approach to optimize large language models by selectively removing non-critical coupling structures based on gradient information to achieve efficient compression while preserving functionality and task agnosticism. Gholami et al. \cite{gholami2023can} demonstrate that weight pruning can be used as an optimisation strategy for the Transfer architecture, proving that judicious pruning can significantly reduce model size without sacrificing performance, thus contributing to bridging the gap between model efficiency and performance.

\subsection{Prompt optimization}
\par Prompt tuning, often referred to as optimization for prompt tuning, is a method used to fine-tune large language model (LLMs) prompts. This methodology does not need access to the underlying model parameters and gradients. This strategy is especially beneficial for closed-source models that have limited access. Prompt optimisation enhances the efficacy of model creation in fewer or zero-shot scenarios by fine-tuning the input prompts. Optimization algorithms (OAs), renowned for their adaptability and efficiency in situations where the internal workings are unknown, are used to discover prompts that improve job performance just by using the outcomes of LLMs reasoning. Prompt tuning may be categorized into two distinct types: continuous and discrete. Continuous prompt tuning employs continuous optimisation algorithms, such as CMAES \cite{hansen2003reducing}, to improve the quality of embedding prompts \cite{chao2024match}. This process involves techniques like partitioning and subspace decomposition to enhance the embedding space. Conversely, discrete prompt tuning uses discrete optimization algorithms to explore the prompt space directly. It utilizes specialized genetic operators to adjust prompts in order to address the problem of combinatorial explosions in the search space. Table \ref{optimization-prompts} summarises the main methods of optimizing prompts

\begin{table*}[tbh]
\caption{Summary of optimization prompts} \label{optimization-prompts}
\centering
\begin{tabular} {p{2.6cm} p{6.4cm} p{3.4cm} p{3.4cm}}
 \toprule[1.5pt]
\textbf{Algorithms} & \textbf{Main tasks}                                                                                & \textbf{Types}        & \textbf{Using OAs}                    \\ \hline
BBT \cite{sun2022black}                 & Optimising large pre-trained language models published as a service                                & Blackbox optimization & Derivative-Free Optimization, CMAES    \\ \hline
BBTv2 \cite{sun2022bbtv2}               & Adjustment of cues by gradient-independent methods                                                 & Blackbox optimization & CMAES                                 \\ \hline
BBT-VLMS \cite{yu2023black}            & Black-box optimization of visual language models                                                   & Blackbox optimization & Derivative-Free Optimization,CMAES    \\ \hline
Textual inversion \cite{fei2023gradient}   & Optimizing the embedded representation of specific concepts in text                                & Continuous            & Iterative Evolutionary Strategy,CMAES \\ \hline
Clip-tuning \cite{chai2022clip}        & Enabling optimisation of prompts without model weight access rights                                & Continuous            & Derivative-free optimization, CMAES   \\ \hline
RGLF \cite{shen2023reliable}               & Effective hint tuning for LLMs in resource-constrained black-box API settings                      & Continuous            & CMAES                                 \\ \hline
GrIPS \cite{prasad2022grips}          & Prompts for Improving Large Language Models                                                        & Discrete              & Edit-based Search, Genetic Algorithms  \\ \hline
GPS \cite{zhao2023genetic}                & Automatically search for high-performance prompts to optimize model performance for specific tasks & Discrete              & Genetic Algorithm                     \\ \hline
Plum \cite{pan2023plum}               & Optimizing and customising large pre-trained language models                                       & Discrete              & Metaheuristics                        \\ 
\toprule[1.5pt]
\end{tabular}
\end{table*}

\subsubsection{Continuous prompt optimization}
\par Continuous prompt tuning is used to optimize the performance of LLMs and is often employed to tune the embedding of prompts. The embedding vectors of the prompts are iteratively tuned to maximize the performance of the model on a particular task, thus improving the quality and performance of the model generation \cite{chao2024match}. Continuous prompt tuning typically explores different strategies and techniques, such as stochastic embedding, subspace decomposition, and knowledge distillation, in order to improve the embedding quality and the search performance for the optimization of large models. For example, Sun et al. \cite{sun2022bbtv2} presented BBTv2, an improved version of Black-Box Tuning, which uses a divide-and-conquer gradient-free algorithm to optimize prompts at different layers of pre-trained models, achieving comparable performance under few-shot settings. Fei et al. \cite{fei2023gradient} introduced a gradient-free framework for optimizing continuous textual inversion in an iterative evolutionary strategy. It accelerates the optimization process with minimal performance loss and compares performance with gradient-based models with variant GPU/CPU platforms. Pryzant et al. \cite{pryzant2023automatic} proposed Automatic Prompt Optimisation (APO) using numerical gradient descent techniques to automatically improve the prompts of Large Language Models (LLMs), obtaining significant performance gains in various NLP tasks and jailbreak detection. Zheng et al. \cite{zheng2023black} Black-box prompt optimization using subspace learning (BSL) enhances the versatility of prompt optimization across tasks and LLMs by identifying common subspaces through meta-learning, ensuring competitiveness across a variety of downstream tasks.
\par Recently, some researchers have used techniques such as knowledge distillation, variational reasoning, and federated learning to improve search efficiency, generalization, and security. For example, Shen et al. \cite{shen2023reliable} presented techniques for adapting large pre-trained language models (PLMs) to downstream tasks using only black-box API access, achieving competitive performance with gradient-based methods while also considering predictive uncertainty in prompts. Sun et al. \cite{sun2023make} present a set of techniques for improving the efficiency and performance of black-box optimization (BBT-RGB) for tuning large language models without access to gradient and hidden representations, demonstrating its effectiveness in a variety of natural language understanding tasks. Han et al. \cite{han2023gradient} proposed GDFO, which ensembles gradient descent and derivative-free optimization for optimising task-specific successive prompts of large pre-trained language models in black-box tuning scenarios, obtaining significant performance gains over previous state-of-the-art approaches. Sun et al. \cite{sun2023fedbpt} proposed FedBPT for federated black-box prompt tuning, a framework for efficient and privacy-preserving fine-tuning of pre-trained language models through collaborative prompt optimization without access to model parameters, reducing communication and memory costs while maintaining competitive performance. Chai et al. \cite{chai2022clip} introduced the Clip-Tuning technique to enhance search efficiency and offer more detailed and diverse evaluation feedback throughout the black-box tuning procedure. Clip-tuning differs from utilizing a single random projection matrix to reduce dimensionality in BBT. Instead, it utilizes pre-trained sampling on dropout models during inference. This process generates many subnetworks that act as predictive projections of samples in the original high-dimensional space. The search technique achieves faster convergence to the ideal solution by aggregating rewards from predictions made by several subnetworks. 

\par Continuous prompt tuning is a method for optimising the performance of large language models where prompts are represented as continuous vectors that exist in a continuous embedding space. These vectors are optimized in a continuous embedding space. The search space is continuous and can be differentiated, allowing the use of gradient-based optimization techniques. Continuous prompt tuning is suitable for black-box optimization scenarios where the internal structure of the model and gradient information are not accessible. However, this approach may require more computational resources due to the complex mathematical operations and gradient calculations involved.

\subsubsection{Discrete prompt optimization}
\par Discrete prompt optimization is a method for finding optimal prompts in a pre-trained language model, where the prompts are represented as discrete text sequences. These methods typically use genetic algorithms, particle swarm optimization, or other heuristic-based search methods to search in a discrete prompt space to find the optimal prompt sequence \cite{wu2024evolutionary,chao2024match}. Unlike continuous prompt tuning, discrete prompt optimization focuses more on tuning prompts at the level of text sequences and is suitable for tasks based on text sequences, such as text generation or classification. Before the emergence of large language models, researchers have been inclined to investigate the application of optimization methods to enhance the performance of pre-trained language models. For instance, Greedy Teaching Prompt Search (GrlPS) \cite{prasad2022grips} uses a stepwise search approach without gradients, whereas Genetic Prompt Search (GPS) \cite{xu2022gps} is grounded in the ideas of genetic algorithms. The majority of these research employ evolutionary algorithms (EAs) as the primary search engine, whereas the language model is tasked with generating and assessing potential prompts \cite{zhao2023genetic}. Within the discrete prompt space, specialized genetic operators are employed to fine-tune heuristics and directly identify the most optimal prompts. This, in turn, enhances the quality of the model's response to a given task. 

\par Typically, these studies employ optimization algorithms as a search framework, where Large Language Models (LLMs) are used to generate and evaluate prompts. Nevertheless, it is important to acknowledge that this research mainly concentrates on particular rapid engineering situations and has restricted breadth. To fully harness the potential of discrete optimization in black-box prompt optimization, it is necessary to tackle the issue of combinatorial explosion in discrete search spaces. For example, Zhou et al. \cite{zhou2023survival} proposed a simple black-box search method called ClaPS, which achieves state-of-the-art performance on a variety of tasks and LLMs while significantly reducing the search cost by clustering and pruning the search space to focus on key prompting tokens that affect LLM prediction. Yu et al. \cite{yu2023black} proposed a black-box prompt tuning framework for visual-verbal models, which optimizes visual and verbal prompts in an intrinsic parameter subspace through an evolutionary strategy, enabling task-relevant prompt learning without back-propagation. Pan et al. \cite{pan2023plum} introduced meta-heuristic algorithms as a generic prompt learning method, and demonstrated their effectiveness in black-box prompt learning and Chain-of-Thought prompt tuning by testing six typical methods and were able to discover more understandable prompts, opening up more possibilities for prompt optimisation. Lapid et al. \cite{lapid2023open} proposed a method for attacking large language models using genetic algorithms to reveal the vulnerability of the models to malicious manipulation and to provide a diagnostic tool for assessing and enhancing the consistency of language models with human intentions. Guo et al. \cite{guo2024connecting} presented EvoPrompt, a discrete prompt optimization framework for the automatic optimization of LLMs prompts using evolutionary algorithms. By linking LLMs and EAs, the method achieved significant performance improvements over manually designed prompts and existing automatic prompt generation methods on 31 datasets, demonstrating the potential of combining LLMs and EAs. Pinna et al. \cite{pinna2024enhancing} present a method for improving the generation of code for large language models using genetic improvement techniques, which significantly improves the quality of the generated code through user-supplied test cases, demonstrating the potential of combining LLM with evolutionary techniques.

\par Generally, discrete prompt optimisation is a method to optimise the performance of large language models in a black-box environment, which searches for optimal prompt words or phrases in a discrete prompt space through techniques such as genetic algorithms, heuristic search, and clustering pruning without the need for internal gradient information of the model. The advantages include effective performance improvement without relying on the internal information of the model in a black-box environment, and adaptability to tasks with a small number of samples or zero samples, while the disadvantages include the possibility of facing a huge search space, the tendency to fall into local optimums, the sensitivity of hyper-parameters, the limited ability of generalisation, the difficulty of interpreting the results, and the high dependence on the choice of evaluation metrics.

\subsubsection{Blackbox optimization prompt tuning}
\par Black-box optimization of large models refers to the process of optimizing and tuning large pre-trained models (e.g., large language models) within a black-box optimization framework. Compared to traditional black-box optimization, black-box optimization of large models is more challenging because the complexity of large pre-trained models and a large number of parameters makes the optimization process more complex and time-consuming. In black-box optimization of large models, optimization algorithms typically optimize the performance of pre-trained models by interacting with them and adjusting their inputs or parameters step-by-step without having direct access to the internal structure or parameters of the model. 
\par In recent years, a number of researchers have focused on the application of black-box optimization to large-scale language models (LLMs) and visual-linguistic models, proposing a variety of methods to optimize model performance without accessing the model's internal parameters or gradients. Yu et al. \cite{yu2023language} optimize a visual language model using a dialogue-feedback-based approach. Guo et al. \cite{guo2023black} introduced the collaborative black-box tuning (CBBT) technique. Sun et al. \cite{sun2022black} develop the a black-box tuning framework for Language Models as a Service (LMaaS). Diao et al. \cite{diao2022black} proposed a black-box discrete prompt learning (BDPL) algorithm. And the work of Yu et al. \cite{yu2023black} introduces a black-box prompt tuning framework for visual language models. These studies demonstrate that in black-box scenarios where model weights cannot be directly modified, external prompt learning and optimization are used to effectively improve model performance in image classification, text-to-image generation, and adaptation to different downstream tasks.

\subsection{Self-Tuning optimization}
\par Compared to the initialization method under the EvoPrompt framework, which relies on hand-prompted optimization. In recent years some researchers have proposed automated prompting methods. Use as a gene operator in EAs to automatically create high-quality prompts for yourself and others. Table \ref{self-Tuning} summarises the main Self-Tuning optimization methods. For example, Singh et al. \cite{singh2023explaining} applied an interpretable auto-prompt (iPrompt) to generate a natural language string that explains the data. Fernando et al. \cite{fernando2023promptbreeder} propose a self-improving mechanism for PromptBreder that evolves and adapts cues for different domains, outperforming existing strategies on arithmetic, common-sense reasoning, and hate speech classification tasks. Pryzant et al. \cite{Pryzant2023AutomaticPO} proposed a simple and non-parametric solution, Automated Prompt Optimisation (APO), which automatically performs fast improvement prompts by using techniques inspired by numerical gradient descent. Li et al. \cite{li2023spell} proposed SPELL, a black-box evolutionary algorithm that uses a large language model to automatically optimize text style cues, demonstrating rapid improvements for a variety of text tasks. 
\par Furthermore, LLMs can serve as a flexible prompt selector for jobs that are not inside the domain it was trained on. Self-tuning can operate inside a versatile language domain without being dependent on parameter updates \cite{chao2024match}. For example, Zhang et al. \cite{Auto-Instruct} proposed Auto-Instruct, an approach that utilizes the generative power of LLMs to automatically improve the quality of instructions for a variety of tasks, going beyond manually written instructions and existing baselines in a variety of out-of-domain tasks, with significant generalisability to other LLMs.

\begin{table*}[tbh]
\caption{Summary of self-Tuning optimization} \label{self-Tuning}
\centering
\begin{tabular} {p{2.6cm} p{6.4cm} p{3.2cm} p{3.4cm}}
\toprule[1.5pt]
\textbf{Algorithms}         & \textbf{Main tasks}                                                                                                                           & \textbf{Objectives}            & \textbf{Using OAs}                        \\ \hline
SPELL \cite{li2023spell}                       & Combining Evolutionary Algorithms and Text Generation Capabilities of LLMs to Optimise Prompts for LLMs in a Black Box Environment            & Classification accuracy        & Variants based on evolutionary algorithms \\ \hline
Promptbreeder \cite{fernando2023promptbreeder}              & Improves LLM performance on specific tasks by automating the cue search and optimisation process without the need to manually engineer prompts & Performance score              & Genetic Algorithm                         \\ \hline
Interpretable Autoprompting \cite{singh2023explaining} & Iteratively use LLM to generate explanations and reorder them based on their performance when used as prompts                                 & Accuracy and interpretability  & Iterative local search algorithm          \\ \hline
APO \cite{Pryzant2023AutomaticPO}                         & Automatically improves prompts to reduce the heavy trial and error required to write prompts manually                                         & Performance of Initial Prompts & Textual gradient descent                  \\ \hline
Auto-Instruct \cite{Auto-Instruct}              & Automatic generation and optimisation of instructions for LLMs                                                                                & Classification accuracy        & Metaheuristic algorithms                  \\ \toprule[1.5pt]
\end{tabular}
\end{table*}

\subsection{Optimize network architecture search}
\par Prompt-based optimization tools improve the quality of model output by optimizing the input format. Another approach known as LLM Network Architecture Search (NAS) focuses on directly optimizing the architecture of LLM models, and in the context of Large Language Models (LLMs), NAS can take a different form by optimizing the architecture of the model directly rather than by tuning the parameters of the model. Table \ref{network-architecture} summarises the main optimized network architecture search search methods

\par As the complexity of neural network models increases, manually designing efficient network architectures becomes time-consuming and challenging. NAS eases the burden on researchers by automating the design process, allowing efficient exploration of the vast search space to discover more efficient, generalized and less resource-intensive model architectures \cite{liu2021survey,wu2024evolutionary, zhou2021survey}. Previously NAS was optimized by simulating the process of natural selection. It involves the steps of randomly generating an initial population, selection, crossover (or recombination, as it is called), and mutation until termination conditions are met \cite{elsken2019neural}. With the development of deep learning techniques and the increase of arithmetic power, the NAS field is also exploring new optimization strategies to optimize large models. With the increase of computational resources and the proposal of new algorithms, the efficiency and effectiveness of NAS have been significantly improved. Nasir et al. \cite{nasir2023llmatic} proposed a new NAS algorithm that effectively combines the advantages of LLMs and Quality Diversity (QD) algorithms to automate the search and discovery of high-performance neural network architectures. So et al. \cite{so2019evolved} proposed an evolutionary Transformer discovered through evolutionary architectural search in multilingual tasks superior Transformer that achieves better performance with fewer parameters and maintains high quality even at smaller sizes.

\par More sophisticated and effective search strategies have been proposed by researchers in recent years to improve the performance of large models. For example, Gao et al. \cite{gao2022autobert} proposed an automatic method (AutoBERT-Zero) for discovering the backbone structure of a general-purpose language model (LLM) using a well-designed search space and an operation-first evolutionary strategy, as well as a two-branch weight-sharing training strategy, to improve search efficiency and performance. Ganesan et al. \cite{ganesan2021supershaper} perform task-independent pre-training of BERT models while generating differently shaped sub-networks by varying the hidden dimensions in the Transformer layer. Rather than optimizing for a specific task, it generates a series of different-sized models by varying the hidden dimensions of the network, which can be fine-tuned for various downstream tasks. Yin et al. \cite{yin2021autotinybert} proposed the use of one-shot Neural Architecture Search (one-shot NAS) to automatically search for architectural hyperparameters. A large SuperPLM is obtained through one-shot learning, which can be used as a proxy for all potential sub-architectures. An evolutionary algorithm is also used to search for the best architectures on the SuperPLM, and then the corresponding sub-models are extracted based on these architectures and further trained. Javaheripi et al. \cite{javaheripi2022litetransformersearch} proposed a no-training Neural Architecture Search (NAS) algorithm for finding Transformer architectures that have an optimal balance between task performance (perplexity) and hardware constraints (e.g., peak memory usage and latency). Zhou et al. \cite{zhou2024training} proposed a Transformer architecture search method called T-Razor, which uses zero-cost agent-guided evolution to improve the search efficiency and evaluates and ranks Transformers by introducing metrics such as synaptic diversity and synaptic saliency to efficiently find optimized architectures in the Transformer search space. Klein et al. \ cites {klein2023structural} proposed Neural Architecture Search (NAS) based on weight sharing as a structural pruning method for finding the optimal balance between optimization efficiency and generalization performance to achieve compression of large language models (LLMs) in order to reduce the model size and inference latency.

\par Overall, the main advantage of NAS for optimising large models is its ability to automate the exploration and discovery of efficient network architectures for specific tasks, significantly improving model performance while reducing manual design and tuning efforts. With intelligent search strategies, NAS helps save computational resources and time. However, this approach also faces challenges, including the large search space, the possibility of falling into local optimal solutions, and the large amount of computational resources required in the initial search and training phases. In addition, the selection and tuning of optimization algorithms require expertise, and the generalization ability of the network architecture obtained from the search still needs further validation. Future research may focus on improving the search efficiency, reducing the computational cost, and enhancing the generalisability and adaptability of the model.

\begin{table*}[tbh]
\caption{Summary of OAs-based network architecture for LLMs} \label{network-architecture}
\centering
\begin{tabular} {p{2.6cm} p{6.4cm} p{3.2cm} p{3.4cm}}
\toprule[1.5pt]
\textbf{Algorithms} & \textbf{Main tasks}                                                                                                           & \textbf{Objectives}        & \textbf{Using OAs}             \\ \hline
Autobert-zero \cite{gao2022autobert}       & Automatic exploration of new self-attentive structures and overall efficient pre-trained language model backbone architecture & Classification performance & Operation-Priority NAS         \\ \hline
AutoTinyBERT \cite{yin2021autotinybert}        & Automatic hyperparameter optimization for efficient compression of pre-trained language models                                & Classification performance & One-shot NAS                   \\ \hline
Llmatic \cite{nasir2023llmatic}             & Discover diverse and powerful neural network architectures                                                                    & Network architecture       & Quality-Diversity optimization \\ \hline
SuperShaper \cite{ganesan2021supershaper}         & Discovering networks with effective trade-offs between accuracy and model size                                                & Classification accuracy    & Evolutionary algorithm         \\ \hline
EvoPrompting \cite{chen2024evoprompting}       & Exploring the use of LLMs as generalised adaptive variation and crossover operators for NAS algorithms                        & Classification accuracy    & Evolutionary algorithm         \\ \toprule[1.5pt]
\end{tabular}
\end{table*}

\section{Application of LLMs-based Optimization Algorithms}\label{app}

\begin{figure}[ht]
  \centering
  \centerline{\includegraphics[width=\linewidth]{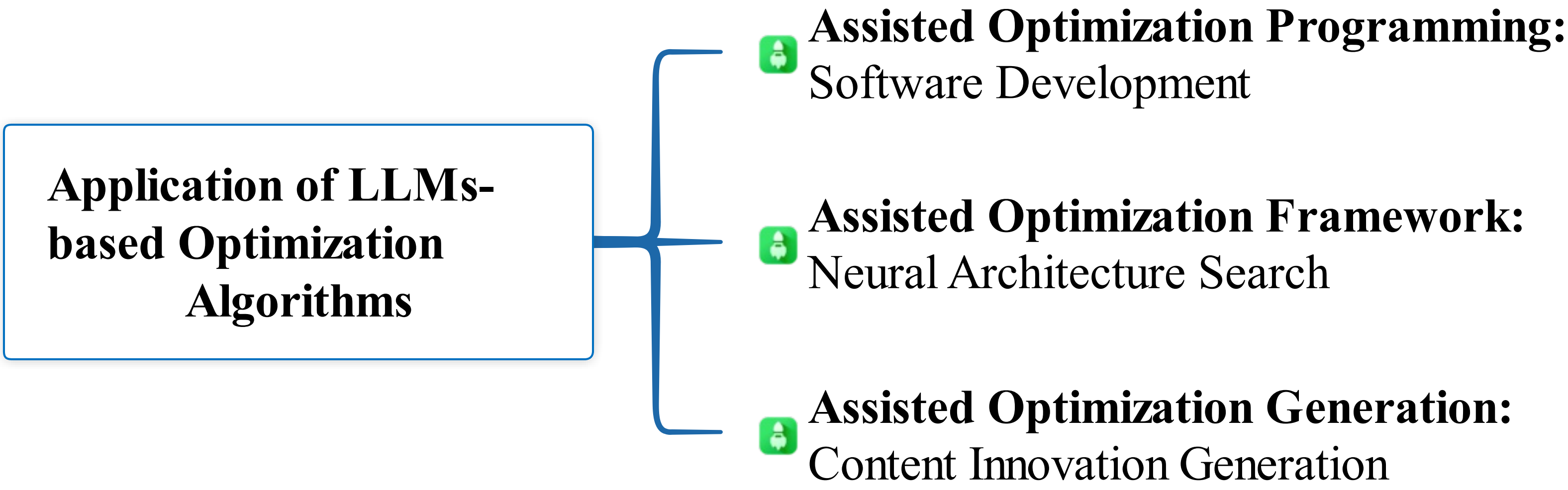}}
\caption{The split of LLMs assist optimization algorithms.}
\label{app_fig}
\end{figure}

As show in Fig.~\ref{app_fig}, optimization algorithms are pivotal in various applications, broadly categorized into software programming, neural architecture search and content generation.
LLM-based optimization algorithms are becoming increasingly important in artificial intelligence, especially in machine learning. They are used for software programming and neural architecture search to help design efficient network architectures. Furthermore, these algorithms are employed as innovative tools in content generation, optimizing the creation process to produce relevant and engaging content. This bifurcation in application highlights the versatility and evolving role of optimization algorithms in addressing both conventional challenges and pioneering technological advancements.

\subsection{Assisted Optimization Programming: Software Programming}
In the wave of artificial intelligence, optimization algorithms based on LLMs have gradually become an important research area to promote code generation and software development. With many parameters and deep learning capabilities, LLMs have shown powerful capabilities in many fields, such as natural language processing, image recognition, etc. Especially in the process of software development, the application of large models can improve the efficiency of code generation and further enhance the model performance through optimization algorithms. By automatically generating code for training models, non-professionals can also easily train efficient machine learning models, greatly reducing the technical threshold and expanding the audience for machine learning technology. Meanwhile, code integration practices in model development, such as static code integration and dynamic code integration, also play a key role in improving the efficiency and quality of software development.

Weyssow et al.~\cite{weyssow2023exploring} explore using Large Language Models (LLMs) for code generation tasks, focusing on Parameter-Efficient Fine-Tuning (PEFT) optimization techniques. The methodology aims to optimize the fine-tuning process of LLMs by updating a small portion of the model's parameters instead of all of them using the PEFT technique to achieve efficient fine-tuning in resource-constrained environments.
Cassano et al.~\cite{cassano2023multipl} present a system called MultiPL-E, which is a system for translating code generation benchmark tests from Python to other programming languages.
Further, Pinna et al.~\cite{pinna2024enhancing} point out the application of automatic code generation based on problem descriptions and that even the most efficient LLMs often fail to generate correct code. Therefore, to address the question of how to enhance code generation based on Large Language Models (LLMs) through a Genetic Improvement (GI) approach, an evolutionary algorithm-based approach is proposed that uses Genetic Improvement (GI) to improve LLM-generated code using a collection of user-supplied test cases. 
Program synthesis (PS), a form of automation, aims to reduce the time and effort required for software development while improving code quality. Although Genetic Programming (GP) is a competing approach to solving the program synthesis problem, it has limitations in evolving syntactically correct and semantically meaningful programs. Tao et al.~\cite{tao2023program} used a combination of Generative Pre-trained Transformers (GPTs) and Grammar-Guided Genetic Programming (G3P) to solve the program synthesis problem. GPTs) and Grammar-Guided Genetic Programming (G3P) to address the program synthesis problem.
The OpenAI team has developed a language model called CodeX~\cite{chen2021evaluating}, which has been fine-tuned using publicly available code on GitHub and investigated for its ability to write Python code. A special production version of Codex is GitHub Copilot, a programming aid.
Brownlee et al.~\cite{brownlee2023enhancing} explored how large language models (LLMs) can be applied to variation operations in Genetic Improvement (GI) to improve the efficiency of the search process.GI is a search-based technique used to improve the non-functional attributes, such as execution time, and functional attributes, such as fixing defects, of existing software.
Ji et al.~\cite{ji2023benchmarking} specifically presents a research overview of the assessment and interpretation of code generation capabilities based on large language models (LLMs), including two main phases: data collection and analysis. In the data collection phase, the prompts' features are quantified by extracting their linguistic features and performance metrics from the generated code. In the analysis phase, causal diagrams are constructed using causal discovery algorithms and further analyzed to identify principles of hint design.

Codex~\cite{chen2021evaluating} was fine-tuned using publicly available GitHub code to enhance its Python coding capabilities. The method evaluated Codex using a new test set, HumanEval, which assesses the functional correctness of programs synthesized from documentation strings.In this dataset, Codex successfully solved 28.8\% of the problems, compared to GPT-3, which solved none, and GPT-J, which solved 11.4\%.
CODAMOSA~\cite{lemieux2023codamosa} integrates a pre-trained Codex with Search-Based Software Testing (SBST) to enhance test case code coverage. SBST generates high-coverage test cases by combining test case generation with mutation for programs under test. However, SBST may face stagnant coverage, meaning it struggles to produce new test cases that increase coverage. When SBST's coverage improvement stagnates, the CODAMOSA algorithm aids in relocating to more advantageous search space areas by using Codex to generate example test cases for functions with lower coverage.
Wu et al.~\cite{wu2023deceptprompt} highlight the advances in code generation by large-scale language models (LLMs) and the associated security risks, particularly the critical vulnerabilities in the generated code. Although some LLM providers have sought to mitigate these issues through human guidance, their efforts have yet to yield robust and reliable code LLMs in practical applications. They introduce the DeceptPrompt algorithm, designed to generate adversarial natural language instructions that prompt code LLMs to produce functionally correct yet vulnerable code. DeceptPrompt employs a systematic, evolution-based algorithm with a fine-grained lossy approach. The algorithm uniquely excels at identifying natural prefixes/suffixes with benign, non-directional semantics and effectively induces code LLMs to generate vulnerable code. This feature enables researchers to conduct near-worst-case red-team tests on these LLMs in real-world scenarios through natural language.

In summary, in software programming, large language models (LLMs) enhance code generation efficiency via optimization algorithms and reduce the complexity of machine learning technology. This simplification allows non-experts to train efficient models, thereby broadening the reach of machine learning.

\subsection{Assisted Optimization Framework: Neural Architecture Search}
Neural Architecture Search (NAS), an important technique for the automatic design of neural networks, is undergoing a transformation driven by combining LLMs and optimization algorithms. With their massive parameters and deep learning capabilities, LLMs show unprecedented potential in handling complex tasks. At the same time, the combination of well-designed optimization algorithms can further accelerate the Neural Architecture Search process, improving the search efficiency and the performance of the resulting models. With the continuous application of big models in NAS, we see their great potential in the automatic search and optimization of neural network structures, which not only greatly saves labor costs but also improves the innovation and diversity of model design.

Nasir et al.~\cite{nasir2023llmatic} present LLMatic, a large model-based Neural Architecture Search (NAS) algorithm that uses two QD archives to search for competitive networks, which combines the code generation capabilities of Large Language Models (LLMs) with the diversity and robustness of Quality Diversity (QD) algorithms.LLMatic utilizes the LLMs to generate new architectural variants and combines the QD algorithms (especially the MAP-Elites algorithm) to discover diverse and robust solutions.
Chen et al.~\cite{chen2024evoprompting} found that an approach combining evolutionary prompt engineering and soft prompt tuning, EvoPrompting, consistently discovers diverse and high-performance models. A method for creating and curating data using evolutionary search to improve in-context prompting examples for LM is presented. While focused on neural architecture design tasks, this approach is equally applicable to LM tasks that rely on in-context learning (ICL) or cue tuning.
Jawahar et al.~\cite{jawahar2023llm} build new uses for Performance Predictors (PP) by using Large Language Models (LLMs) that predict the performance of specific Deep Neural Network (DNN) architectures on downstream tasks. A hybrid search algorithm (HS-NAS) is proposed, which uses LLM-Distill-PP in the initial phase of the search and a baseline predictor for the remainder of the search.HS-NAS reduces the search time by about 50\% with performance comparable to that of the SOTA NAS and sometimes improves latency, GFLOPs, and model size.
Jawahar et al.~\cite{jawahar2023llm} introduced LLM-PP, a precise performance predictor developed using LLM for few-shot prompting. It achieves a mean absolute error (MAE) comparable to the state of the art (SOTA). LLMDistill-PP, developed as a more cost-effective predictor, caters to applications like Neural Architecture Search (NAS) that require numerous predictions. Additionally, the new HS-NAS algorithm is introduced. It leverages the strengths of LLMDistill-PP and the state-of-the-art performance estimator, reducing NAS search times by half and identifying more efficient architectures.

Zheng et al.~\cite{zheng2023can} explored the potential of GPT-4 models for the Neural Architecture Search (NAS) task of designing effective neural network architectures. At the same time, they propose an approach called GPT-4 Enhanced Neural archItectUre Search (GENIUS), which leverages the generative power of GPT-4 as a black-box optimizer to navigate the architectural search space quickly, identify promising candidate architectures, and iteratively refine these candidate architectures to improve performance.
EvoPrompting~\cite{chen2024evoprompting} employs advanced Language Models (LMs) for code-level Neural Architecture Search (NAS). This approach integrates evolutionary prompt engineering with soft-prompt tuning. It aims to iteratively refine contextual prompts and enhance prompt tuning on LMs, thereby boosting their capacity to generate innovative and diverse solutions for complex reasoning tasks.
Radford et al.~\cite{radford2018improving} describe a method to enhance Natural Language Understanding (NLU) using Generative Pre-Training. They show that this approach significantly boosts performance across various NLU tasks by initially pre-training a language model on a vast corpus of unlabeled text, then applying supervised fine-tuning for particular tasks.
Chowdhery et al.~\cite{chowdhery2023palm} proposed PaLM (Pathways Language Model), a large-scale language model, PaLM has demonstrated excellent performance on a variety of Natural Language Processing (NLP) tasks, and PaLM also has very good performance on network structure design, structure search.

In summary, integrating LLMs with optimization algorithms in neural network architecture search (NAS) enhances search efficiency, fosters innovation, diversifies model designs, and opens new avenues for the automated design of complex neural architectures.

\subsection{Assisted Optimization Generation: Content Innovation Generation}
Innovative content generation has become a key driver for developing media, entertainment, arts, and scientific discovery. Applying big artificial intelligence models combined with optimization algorithms is increasingly important in this process. In summary, using optimization algorithms based on large models in innovative content generation is not only about the innovation and diversity of content but also promotes and facilitates scientific and technological innovation development.

Xiao et al.~\cite{xiao2023patterngpt} proposed a pattern-centric text generation framework, PatternGPT, to address the error-prone nature of Large Language Models (LLMs) and the inability to use external knowledge in text generation tasks directly. The framework uses algorithms to search for or generate high-quality patterns based on judgmental criteria. It leverages the pattern extraction capabilities of LLMs to develop a diverse set of structured and formalized patterns, which can help to bring in external knowledge for computation.
Chen et al.~\cite{chen2023mapo} enhance the performance of Large Language Models (LLMs) in language generation tasks through Model-Adaptive Prompt Optimization (MAPO), a prompt optimization method that can be widely applied to various downstream generation tasks.
Similarly, they propose a new paradigm for news summary generation that uses Large Language Models (LLMs) to improve the quality of news summary generation through evolutionary fine-tuning~\cite{xiao2023enhancing}. The method uses LLM to extract multiple structured event patterns from news passages, evolves a population of event patterns via a genetic algorithm, and selects the most adapted event patterns to input into LLM to generate news summaries.
PanGu Drug Model~\cite{lin2022pangu} is a graph-to-sequence asymmetric conditional variational autoencoder designed to improve molecular property representation and performance in drug discovery tasks. The model is inspired by conversions between molecular formulas and structural formulae in the chemistry classroom and can appropriately characterize molecules from both representations.
Liang et al.~\cite{liang2023drugchat} presented a prototype of a DrugChat system designed to provide ChatGPT-like capabilities for drug compound analysis.DrugChat, by combining graph neural networks (GNNs), large language models (LLMs), and adapters, enables users to upload molecular maps of compounds and ask various questions during multiple rounds of interaction. Diagrams and ask different questions in numerous rounds of interaction, which the system then answers.
To break the bottleneck of literate graph technology, Berger et al.~\cite{berger2023stableyolo} proposes the framework of StableYolo, which aims to optimize the image generation quality of large language models (LLMs) by applying evolutionary computation to the Stable Diffusion model while adjusting the prompts and model parameters. The core idea of StableYolo is to improve the image generation quality of photo-realistic styles by combining visual evaluation with multi-objective search. The core concept of StableYolo is to enhance the quality of image generation in photo-realistic style by combining visual evaluation with multi-objective search. The system uses the confidence estimate of the Yolo model as a fitness function and searches for the optimal combination of cue words and model parameters using a Genetic Algorithm (GA).

To explore additional research related to LLMs, including cognitive functions of LLMs, behavior and learning in game-theoretic environments, and Big Five personality traits, Suzuki et al.~\cite{suzuki2024evolutionary} propose a model for the evolution of personality traits based on Large Language Models (LLMs), specifically those related to cooperative behavior. The approach demonstrates how LLMs can enhance the study of human behavioral evolution and is based on evolutionary game theory by using an evolutionary model that assumes that human behavioral choices in game-theoretic situations can be simulated by providing LLMs with high-level psychological and cognitive trait descriptions.
De et al.~\cite{de2023emergence}  explored the phenomenon of the self-organized formation of scale-free networks in social interactions between large language models (LLMs). Scale-free networks are a typical emergent behavior in complex systems, especially in online social media, where users can follow each other and form social networks with specific structural features.
Lu et al.~\cite{lu2023self} propose a novel learning framework, SELF (Self-Evolution with Language Feedback), which aims to continuously enable large-scale language models (LLMs) to improve themselves through self-feedback and self-improvement. The SELF framework is inspired by the human self-driven learning process, which consists of an initial attempt, reflective feedback, and The SELF framework is inspired by the human self-driven learning process, which involves a cycle of initial attempts, reflective feedback, and behavioral improvement to improve the model's capabilities. The ELF framework also enables smaller LLMs to improve themselves, which can be reversed to facilitate the development of larger predictive models. 

In summary, the proposed systems and frameworks, including DrugChat and SELF, illustrate the development of personalized, intelligent tools for analyzing drug compounds, generating news summaries, and mimicking human behaviors. These tools continuously improve their performance through self-learning and feedback mechanisms, enhancing efficiency and accuracy in related fields.

\section{Future Outlook and Research Trends} \label{future}
In the previous sections, we have examined recent advances in the fields of long-term memory models (LLMs) and optimization algorithms (OAs). Nonetheless, there are still many challenges and unresolved issues between these two fields. Therefore, the aim of this section is to explore directions for future research in order to provide scholars with the opportunity to explore new areas beyond the boundaries of current knowledge, to ask new research questions, and to reinvigorate the field.

\textbf{Theoretical Foundations and Methodologies.} Experimental studies have confirmed the effectiveness of combining large-scale language models (LLMs) with OAs in solving small-scale problems \cite{liu2023large,meyerson2023language}. However, the motivation for their interaction has not yet been clarified. To further promote the performance of algorithms, we need to deeply explore the mechanism of mutual reinforcement between LLMs and OAs in theoretical studies and analyze their complementary advantages and potential problems in practical applications in detail through large-scale empirical studies. In addition, it is crucial to conduct in-depth theoretical analyses of algorithms combining LLMs and OAs, which includes evaluating their convergence, time complexity and space complexity. Also, investigating the impact of algorithmic parameter settings on performance, as well as performance guarantees or theoretical limitations of the algorithms on different problem types, are key steps in advancing the algorithms. Further, exploring optimization theory \cite{liu2023pre}, such as clarifying the definition and characterization of the objective function, dealing with constraints, and analyzing the feasible solution space of a problem, will provide a solid theoretical foundation for the design and application of algorithms to achieve better algorithmic performance in solving more complex problems.

\textbf{Automated Intelligent Optimization.} In the optimization context, large language models (LLMs) show significant potential, especially in enhancing the automation and intelligence of optimization algorithms (OAs). Learning from multimodal data during the pre-training phase allows LLMs to understand and generate cross-modal content \cite{wu2023multimodal}. This provides a new search and mutation strategy for OAs when performing cross-modal operations. This capability of LLMs can facilitate OAs in achieving a more efficient global search in multimodal optimization problems. At the same time, as the technology of LLMs continues to advance, it is expected that they will drive the performance of OAs in modeling complex evolutionary mechanisms, especially when dealing with optimization problems with large-scale search spaces \cite{cui2024survey}. However, current research has yet to explore the potential of LLMs in evolutionary optimization, and there remain challenges, such as how to combine LLMs and OAs better and how to handle complex search spaces.

\par In addition, the pre-training of LLMs on large amounts of textual data embeds them with rich domain knowledge, which provides a robust knowledge base for OAs.LLMs can assist OAs in better integrating domain-specific knowledge in the optimization process, thus improving the efficiency of optimization and the quality of solutions. For example, LLMs can generate high-quality initial solutions, improve problem formulation, and provide solution coding and definition of solution spaces. In addition, LLMs can provide guiding principles for algorithm design, enabling EAs to handle complex optimization problems such as multi-objective, discrete and dynamic more effectively \cite{tan2021evolutionary}. With the rapid development of LLMs technology, they are expected to play an even more critical role in the future evolutionary optimization field, driving the field toward higher levels of automation and intelligence.

\textbf{Robustness and Prompt Engineering.} Utilizing optimization techniques is a crucial method for improving the capabilities of LLMs in engineering applications. Common approaches involve utilizing LLMs as optimization operators within EIA frameworks to consistently produce fresh prompt. This technique has consistently shown efficacy and superiority in numerous investigations. Nevertheless, certain obstacles persist. Firstly, it is crucial to pay close attention to the initialization of the optimization process as it will have a substantial impact on the outcomes \cite{chen2024evoprompting,li2023spell}. It is crucial to have cue templates that are generic and customizable in order to provide accurate and valid prompts. Random initialization may not be capable of using existing information, and manual seeding may add bias. In addition, when confronted with issues that contain a significant amount of previous knowledge, the range of possible prompts to consider increases exponentially as the length of the cue and the size of the vocabulary grow. This can result in over fitting or becoming trapped in local optimal solutions. Furthermore, these approaches lack stability and strongly depend on the capabilities of the LLM, rendering them susceptible to stochasticity \cite{li2023spell}. If the LLM lacks the ability to comprehend and efficiently employ the cues, it may undermine the effectiveness of the approach.
\par Further research should strive to tackle these obstacles by creating more resilient techniques. For instance, in the context of initialization, the technique of multisource seeding can be investigated to automatically improve the size and quality of the initial population utilising LLM. When dealing with intricate search spaces, it is essential to develop efficient optimisation algorithms. This may involve combining more comprehensive sets of optimisation operators, using the advantages of different evolutionary algorithms, and utilising adaptive optimisation techniques.

\textbf{Generality and Architecture Search.}
The combined efforts of Large Language Models (LLMs) and OAs have accelerated progress in the field of code generation, leading to notable improvements in downstream applications such as software engineering and OAs design. An commonly used method in this collaboration involves employing LLMs to create large training datasets, and then refining the LLMs using reinforcement learning approaches \cite{lehman2023evolution,chen2023seed}. Nevertheless, this approach has challenges with the variety and quantity of training data, which could result in a failure to cover all possible scenarios. An alternate approach involves utilizing the strong code generation capabilities of LLMs in combination with the powerful search architecture of OAs to continually improve the code generation process. However, this method has difficulties when it comes to generating code for sophisticated algorithmic logic that may require the combined work of numerous code snippets. In order to overcome these obstacles, it is possible to develop a modular strategy that breaks down large activities into smaller, more manageable sub-tasks. An interactive interface might be added to allow users to clearly define the breakdown of tasks \cite{chen2023seed}. This would enable LLMs and OAs to generate code for each sub-task in a coordinated manner.

\par Neural Architecture Search (NAS) is an important application scenario that arises from the combination of LLMs and OAs. Although LLMs have shown remarkable effectiveness in other tasks, they have not been specifically designed for NAS \cite{chen2024evoprompting}. The performance of current LLM models varies significantly when used for NAS tasks, and there is a clear difference between LLM-based approaches and conventional NAS methods in terms of their application area and ability to generalize \cite{ren2021comprehensive}. In order to enhance the overall effectiveness of LLMs and EAs in NAS projects, a comprehensive strategy could be implemented. This involves assessing the effectiveness of various LLM models in NAS tasks, enhancing LLM's NAS skills by incorporating more training data, optimizing the structure of LLMs during the fine-tuning stage, and investigating the utilization of past search knowledge to speed up future searches and provide clearly defined search spaces for LLMs.

\par \textbf{Interdisciplinary Applications and Innovations.} The incorporation of Large Language Models (LLMs) with optimization algorithms (OAs) shows potential in several interdisciplinary domains, providing a powerful synergy to stimulate innovation and improve performance in intricate jobs.

\par In the realm of computational creativity and generative design, LLMs are adept at generating creative content, such as artwork, music, and literary pieces. The collaboration with OA brings methods of variation and selection, which can promote creative diversity and ignite innovation. This collaborative approach can result in the production of unique and groundbreaking artistic and design works, thereby promoting innovation and fostering the growth of creativity. Within the domain of robotics, intelligent and adaptable robot systems can be produced through the collaboration of OAs, which have the ability to refine control strategies and action sequences, and LLMs, which are capable of generating instructive dialogues and task-oriented directives \cite{wu2024evolutionary}. These systems have enhanced capabilities to adjust to various tasks and participate in complex interactions with humans, enhancing collaboration between humans and robots and establishing the foundation for advanced robotic applications.

\par Moreover, in the field of drug design, the ability of LLMs to produce new chemical structures, combined with the multi-objective optimization capabilities of OAs, can accelerate the process of discovering new drugs. This comprehensive technique has the capability to recognize drug candidates with higher potential, so decreasing the time and expenses linked to conventional trial-and-error procedures and promoting progress in pharmaceutical research and development. The combination of LLMs and OAs offers a versatile instrument that has the ability to transform various fields by offering inventive solutions and improving efficiency in problem-solving. As research explores the joint capabilities of LLMs and OAs, it is expected that more significant advancements and innovative uses will arise, revolutionizing industries and expanding the limits of human accomplishment.


\printcredits

\bibliographystyle{cas-model2-names}

\bibliography{refs}

\end{document}